\documentclass[10pt,twocolumn,letterpaper]{article}

\usepackage{iccv}
\usepackage{times}
\usepackage{epsfig}
\usepackage{graphicx}
\usepackage{amsmath}
\usepackage{amssymb}
\usepackage{bm}
\usepackage{tabularx}
\usepackage{caption}

\newcommand{\V}[1]{\mathbf{#1}}

\usepackage[pagebackref=true,breaklinks=true,letterpaper=true,colorlinks,bookmarks=false]{hyperref}

\iccvfinalcopy 

\def\iccvPaperID{9477} 

\ificcvfinal\fi

\begin{document}
\title{Physics-based Human Motion Estimation and Synthesis from Videos}

\author{Kevin Xie$^{1,2}$, Tingwu Wang$^{1,2}$, Umar Iqbal$^{2}$\\ Yunrong Guo$^{2}$, Sanja Fidler$^{1,2}$, Florian Shkurti$^1$\\
$^1$University of Toronto and Vector Institute, $^2$Nvidia\\
{\tt\small kevincxie@cs.toronto.edu}
}
\twocolumn[{%
\renewcommand\twocolumn[1][]{#1}%
\maketitle
\newcommand\hh{3.6cm}
\begin{center}
\vspace{-8mm}
    \captionsetup{type=figure} 
    \includegraphics[height=\hh,trim=370 150 160 250,clip]{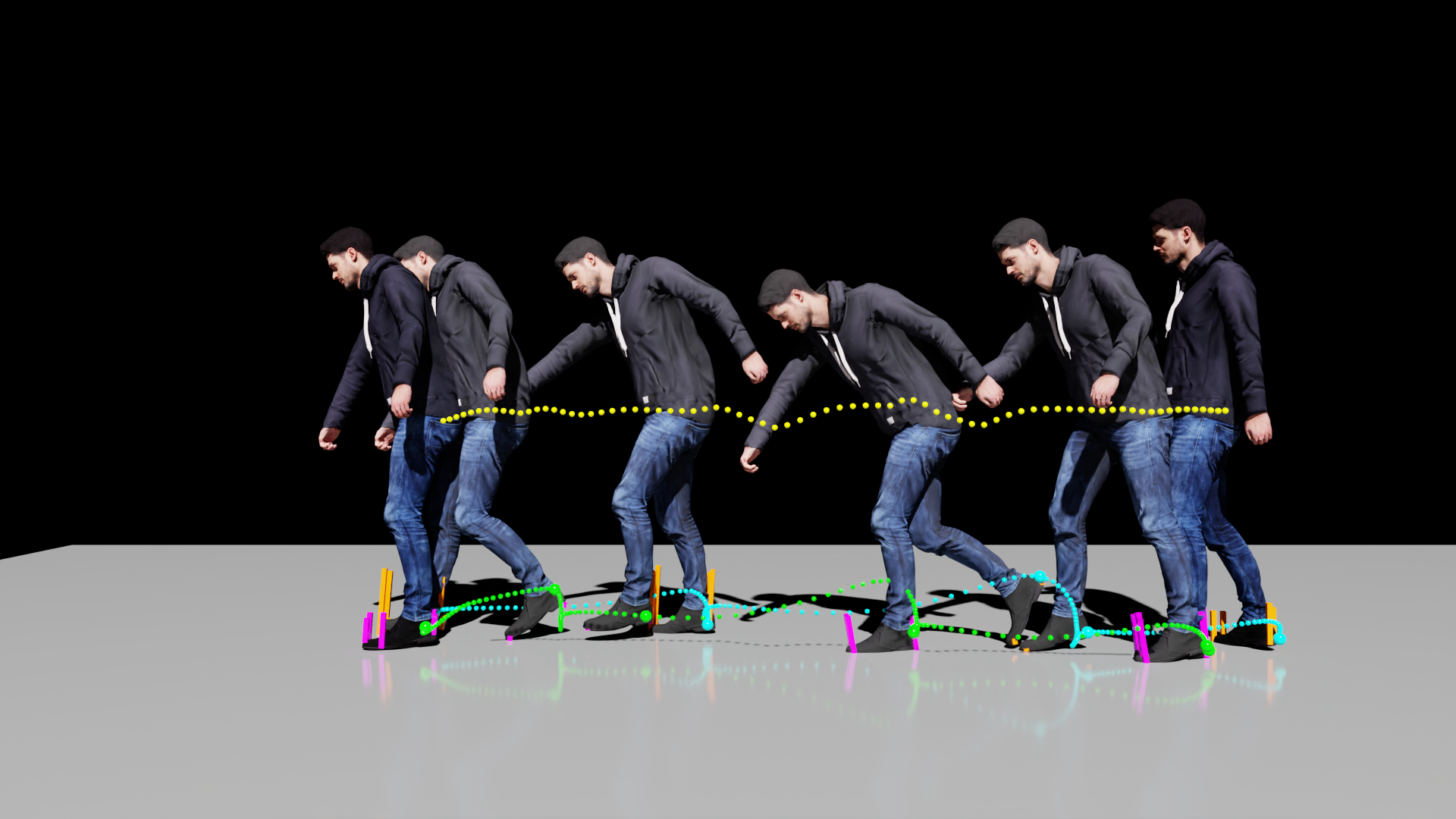}\hspace{0.5mm}\includegraphics[height=\hh,trim=490 30 560 220,clip]{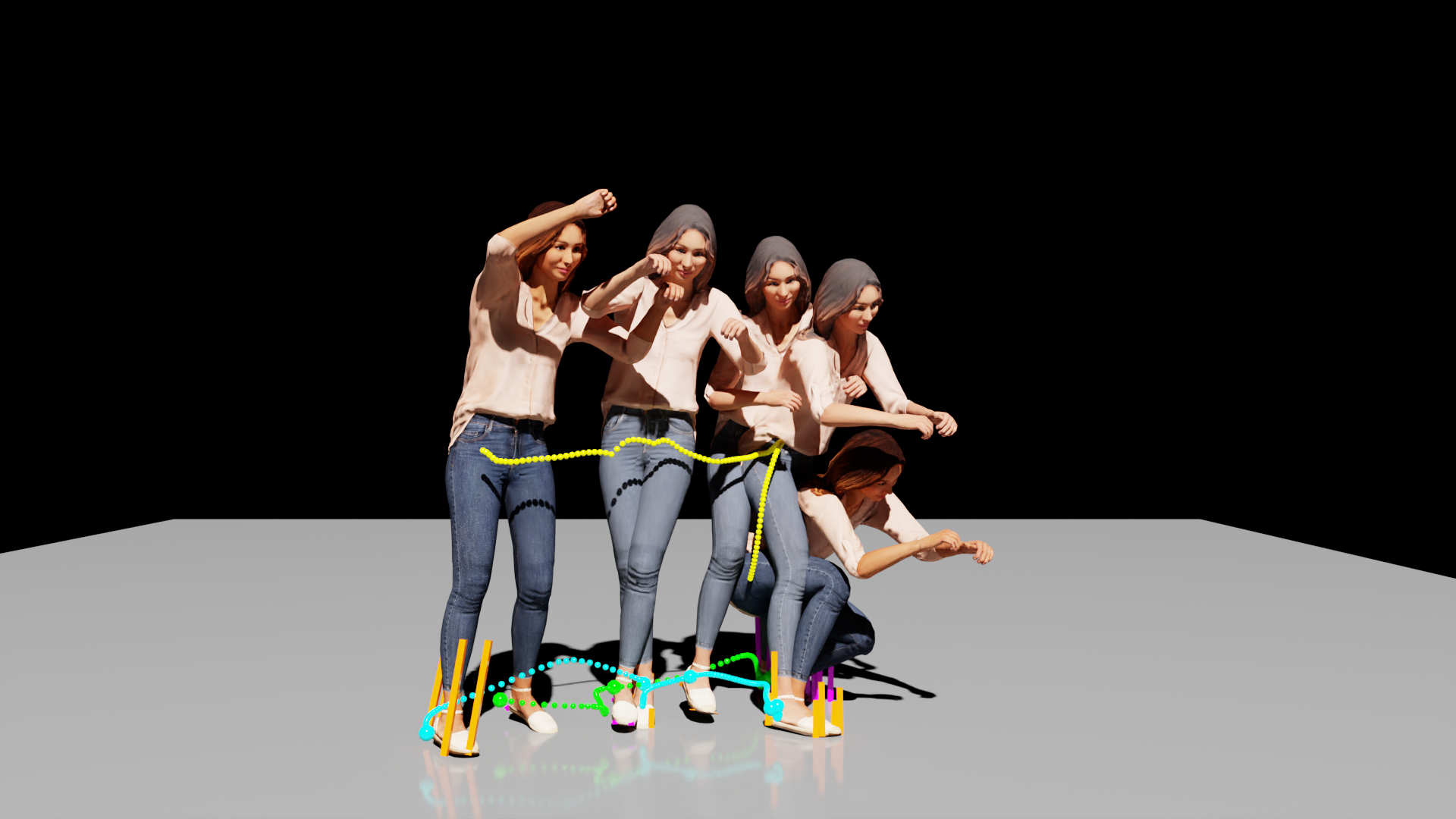}\hspace{0.5mm}\includegraphics[height=\hh,trim=100 10 100 30,clip]{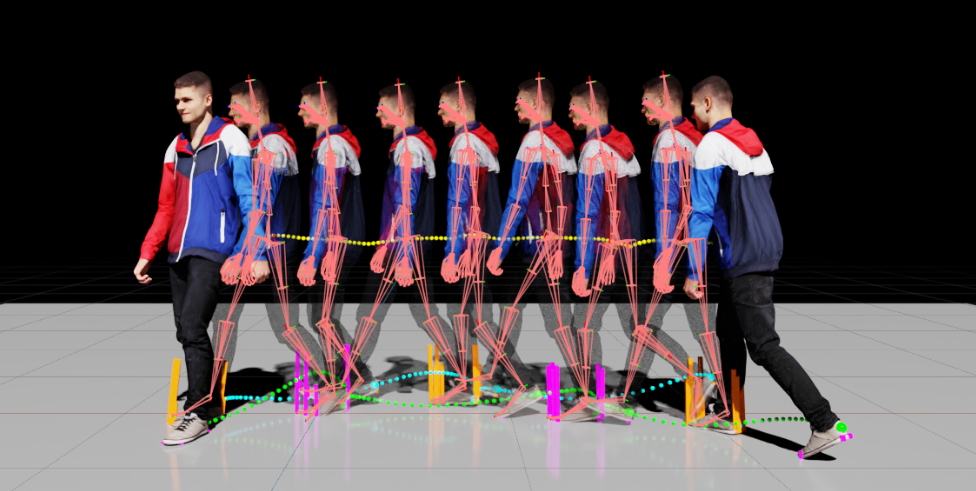}
    \vspace{-6mm}
    \captionof{figure}{We propose a framework to estimate physically correct motions from noisy pose estimations from video. This allows us to train a motion synthesis network directly on video data, removing the need for mocap data used in prior work.}
\end{center}
}]
\begin{abstract}
\vspace{-2mm}
Human motion synthesis is an important problem with applications in graphics, gaming and simulation environments for robotics. Existing methods require accurate motion capture data for training, which is costly to obtain. Instead, we propose a framework for training generative models of physically plausible human motion directly from monocular RGB videos, which are much more widely available.
At the core of our method is a novel optimization formulation that corrects imperfect image-based pose estimations by enforcing physics constraints and reasons about contacts in a differentiable way.
This optimization yields corrected 3D poses and motions, as well as their corresponding contact forces. Results show that our physically-corrected motions significantly outperform prior work on pose estimation. 
We can then use these to train a generative model to synthesize future motion.
We demonstrate both qualitatively and quantitatively improved motion estimation, synthesis quality and physical plausibility achieved by our method on the Human3.6m dataset~\cite{h36m_pami} as compared to prior kinematic and physics-based methods. By enabling learning of motion synthesis from video, our method paves the way for large-scale, realistic and diverse motion synthesis. Project page: \url{https://nv-tlabs.github.io/publication/iccv_2021_physics/}
\end{abstract}
\vspace{-4mm}
\section{Introduction}
Given videos of human motion, how can we infer the 3D trajectory of the body's structure and use it to generate new, plausible movements that obey physics constraints? Addressing the intricacies of this question opens up an array of possibilities for high-fidelity character animation and motion synthesis, informed by real-world motion. This would benefit games, pedestrian simulation~\cite{yang2020recovering} in testing environments for self-driving cars, realistic long-horizon predictions for model-based control and reinforcement learning, as well as physics-based visual tracking.            

The vast majority of existing approaches in learning-based human motion synthesis ~\cite{aksan2019structured, yan2021mojo, motion_vae, motion_fields_zoran, motion_matching_clavet} rely on large-scale motion capture observations, such as AMASS~\cite{AMASS:ICCV:2019}, which are typically costly and time-consuming to acquire, logistically challenging, and most often limited to recordings in indoor environments.  These factors form a bottleneck that hinders the collection of high-quality human motion data, particularly in settings where there is interaction among multiple people or interaction with a number of stationary and moving objects in the scene. The recorded motions typically also lack realism and diversity as they are acquired by acting out a set of pre-defined motions. In addition to this issue, many time-series models trained on motion capture data make predictions that are oblivious to the physics constraints of motion and contact, often leading to inaccurate, jerky, and implausible motion.      

In this paper we entirely forego reliance on motion capture and aim to train physically plausible human motion synthesis directly from monocular RGB videos. We propose a framework that refines noisy image-based pose estimates by enforcing physics constraints through contact invariant optimization~\cite{mordatch2012discovery,mordatch2012contact}, including computation of contact forces.

We then use the results of the refinement to train a time-series generative model that synthesizes both future motion and contact forces. Our contributions are:
\begin{itemize}
    \item We introduce a smooth contact loss function to perform physics-based refinement of pose estimates, eschewing the need for separately trained contact detectors or nonlinear programming solvers. 
    \item We demonstrate that when visual pose estimation is combined with our physics-based optimization, even without access to motion capture datasets, it is sufficient to train motion synthesis models that approach the quality of motion capture prediction models.
\end{itemize}

We validate our method on the Human3.6m dataset~\cite{h36m_pami}, and demonstrate both qualitatively and quantitatively the improved motion synthesis quality and physical plausibility achieved by our method, compared to prior work on learning-based motion prediction models, such as PhysCap~\cite{PhysCapTOG2020}, HMR~\cite{hmrKanazawa18}, HMMR~\cite{zhang2019phd}, and VIBE~\cite{kocabas2019vibe}. 

\section{Related Works}
We organize the rich existing literature on motion synthesis across two axes: (a) kinematic vs. physics-based methods, and (b) imitation learning vs. model-based control and reinforcement learning.  Table~\ref{tab:relwork} provides a summary of the most relevant works. 

\subsection{Kinematic Motion Synthesis}

Kinematic motion synthesis models make predictions without necessarily satisfying physics constraints. \emph{Non-parametric methods} in this category attempt to blend motion clips and concatenate them into a coherent trajectory. Examples of this type of work include motion matching~\cite{motion_matching_clavet} and the use of motion graphs~\cite{motion_graphs, motion_graphs_reitsma} and motion fields~\cite{motion_fields_zoran} in character animation. 

\emph{Parametric kinematic methods}, on the other hand, rely on pose predictions made by a time-series generative model, typically a neural network. After training, the example motions are not used for prediction anymore, in contrast to non-parametric approaches. To maintain consistency in the predicted motion many papers make use of motion generation via recurrent neural networks (RNN)~\cite{fragkiadaki, Martinez_2017_CVPR, yan2018mt, zhou2018autoconditioned, Ghosh2017LearningHM, starke_local_motion_phases}, variational autoencoders for time-series data~\cite{motion_vae, Habibie2017ARV, yuan2020dlow}, autoregressive models~\cite{moglow, starke_neural_state_machine}, transformers~\cite{Li2020LearningTG}, or by explicitly maintaining a memory bank of past motions. 

\subsection{Physics-Based Motion Synthesis}
Physics-based animation methods make motion predictions that satisfy the body dynamics and are informed by physics constraints~\cite{drecon_clavet}, often including contacts, which adds to the realism of the generated movement. Seminal work in \emph{contact-invariant optimization}~\cite{mordatch2012discovery,mordatch2012contact} introduced soft inverse dynamics constraints to optimize center-of-mass trajectories as well as contact forces without requiring explicit planning of contact locations. In~\cite{interactive_cio} and in~\cite{manocha_2019}, it was shown that this framework could be sped up and also be used in a setting where target velocities are selected interactively.    

In addition to imposing soft physics constraints, recent reinforcement learning controllers have been used for motion synthesis with \emph{hard physical constraints}~\cite{liu2010sampling,liu2015improving}. These approaches leverage model-based sampling planning to generate physically correct motions which corrects pose estimation errors and model mismatch. Injecting hard physics constraints for dynamics and contact has been a fruitful approach in trajectory optimization for humanoid robots~\cite{righetti_variable_mpc}, which typically makes use of nonlinear programming solvers and mixed integer-quadratic programs. Incorporating example motions and training data into these optimization frameworks, however, is challenging. So is generating diverse motions. Moreover, execution times of these frameworks are typically not suitable for real-time operation.

   \begin{table}[t!]
\small
  \addtolength{\tabcolsep}{-2.8pt}
\begin{tabular}{l|ccc}
                                            & Input Modality & Physics        &    Functionality \\ \hline
DLow~\cite{yuan2020dlow}                    & mocap       &                   & synthesis \\
RFC~\cite{yuan2020rfc}                      & mocap       &  $\thicksim$      & synthesis \\
MOJO~\cite{yan2021mojo}                     & mocap       &                   & synthesis              \\
PhysCap~\cite{PhysCapTOG2020}               & video       & \checkmark        &  pose est.          \\
Rempe~\etal~\cite{RempeContactDynamics2020} & video & \checkmark              &  pose est.          \\
Ours                                        & video & \checkmark & pose est., synthesis \\ \hline
\end{tabular}
\vspace{-2.5mm}
\caption{\small Comparison of features of different related works. RFC uses a physics simulator but does not use proper contact dynamics.}
\label{tab:relwork}
\vspace{-4.5mm}
\end{table}

To balance the fidelity of dynamics with the cost of computation time, simplified physics models such as centroidal dynamics models, or models that enforce soft-dynamics constraints, have been commonly used in literature. For example,~\cite{winkler2018gait,kwon2020fast} use centroidal dynamics to fine-tune character motion from physically incorrect motion templates.

Model-free reinforcement learning approaches, which do not assume known or learned dynamics,
are gaining popularity due to their flexibility, efficiency in high-dimensional motion synthesis that tracks realistic reference motion. In DeepMimic~\cite{peng2018deepmimic}, model-free controllers are trained to output torques to follow the reference motion. 
DeepMimic is able to physically correctly reproduce a large variety of motion skills.
However it takes hours or days just to reproduce one motion.
Since then, efforts have been made to extend model-free controllers.
In~\cite{won2020scalable, wang2020unicon}, by improving the capacity of neural networks, 
controllers can now master all the skills in a large motion dataset without having to retrain for each motion as in DeepMimic.

\vspace{-1mm}
\subsection{Kinematic and Physics-Based Pose Estimation}
Purely kinematic approaches for 3D pose and shape~\cite{texmesh, yan2021mojo} estimation from video, such as HMMR~\cite{humanMotionKanazawa19}, VIBE~\cite{kocabas2019vibe}, and XNect~\cite{xnect}, predict past and future motion, without incorporating physics constraints.
Physics constraints, however, can act as a regularizer, adding temporal consistency to the estimated 3D motion. Both motion capture and human video data have been used as observations in pose estimation, with the latter modality leading to an ill-posed problem. For example, PhysCap~\cite{PhysCapTOG2020} achieves physically plausible real-time human motion estimation in 3D from videos, including modeling of contacts and prediction of their locations, which leads to minimal foot-to-floor penetration. ~\cite{RempeContactDynamics2020} also models hard contact constraints, which cannot be changed after detection. Physics-based visual tracking~\cite{vondrak_2008} provides additional examples of work in this area, including ones that handle contacts~\cite{Li_2019_CVPR, brubaker} as hard constraints during trajectory optimization, as well as entire meshes~\cite{liu20204d}.

Our main difference from these works is that by using our proposed soft contact penalty, contact events can form dynamically and softly during optimization. Our method does not need separate contact labelling, and instead of a complex alternating optimization with discrete steps to relabel contacts, it optimizes in two contiguous passes with an off-the-shelf unconstrained LBFGS optimizer.

\section{Method}
\begin{figure}[t!]
    \centering
    \includegraphics[width=1\linewidth]{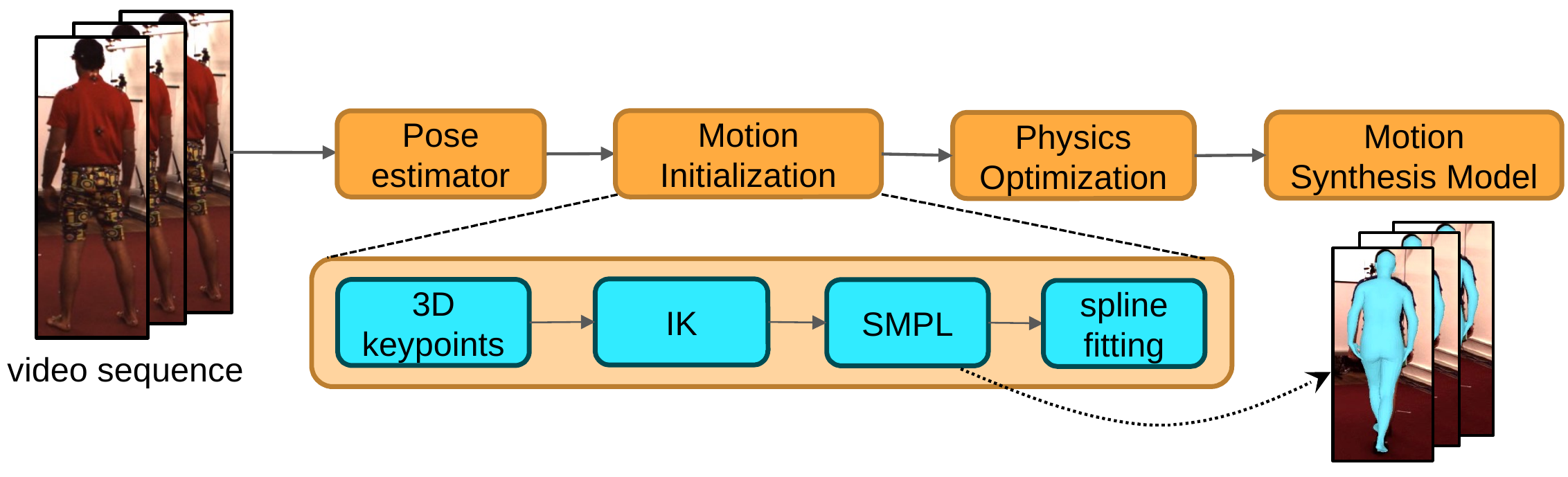}
    \vspace{-7mm}
    \caption{\small{\textbf{Overview of our framework.} A video sequence is processed by a per-frame CNN pose estimator. The 3d and 2d keypoint detections are passed to an inverse kinematics step that forms an initial estimate of the SMPL body model motion using 3D keypoints. We then optimize this initialization with our physics loss and use the produced motions in place of motion capture to train motion synthesis models.}}
    \label{fig:framework}
     \vspace{-2mm}
\end{figure}

An overview of our proposed framework for learning motion synthesis from videos can be seen Fig~\ref{fig:framework}. It consists of four steps: {\bf 1)} Given an unlabeled video, we estimate the positions of 2D and 3D body joints at each video frame using a monocular pose estimation model~\cite{iqbal2020learning}. {\bf 2)} We then transform the 3D body joints at each frame to relative body-part rotations of the parametric body model, SMPL~\cite{SMPL:2015}, using inverse-kinematics~\cite{iqbal2021kama, li2021hybrik}. {\bf 3)} We then refine the initial motion estimates using our proposed physics-based optimization which results in physically plausible and temporal coherent motion for the entire video. {\bf 4)} We process all available videos with aforementioned steps, and subsequently use the resulting motions to train our motion synthesis model. In the following we detail each step. 

\subsection{3D Pose Estimation}
Given an unlabeled RGB video, we estimate the 3D body pose from each frame using a monocular pose estimation model. For this, we chose the method of~\cite{iqbal2020learning,iqbal2018hand} as it provides 3D body pose in absolute camera coordinates. We follow~\cite{iqbal2020learning} and use HRNet-w32~\cite{SunXLW19} as the backbone, and train it on Humans3.6M~\cite{h36m_pami}, 3DPW~\cite{vonMarcard2018} and MSCOCO~\cite{lin2014microsoft} datasets. 
At each frame, it provides the 3D pose $\V{p^{pe}} \in \mathbb{R}^{J\times3}$ and 2D pose $\V{p^{pe,2d}} \in \mathbb{R}^{J\times2}$, where the 3D pose $\V{x}$ is estimated up to a scaling factor. The global scale of the person is approximated using the mean bone length, which obviously is sub-optimal and leads to physically implausible results, \eg. feet penetrating the ground-plane. Since the pose of each frame is estimated independently, we found that the resulting poses contain a large amount of jitter both spatially as well as in terms of scale. 

\subsection{Motion Representation and Initialization}
The 3D positions are not the most optimal representation for effectively modelling the spatial and temporal inter-/intra-part correlations, as slight changes in the depth of the person under the same pose will lead to significantly different 3D positions for all body parts. Hence, we convert the 3D positions to local body-part rotations using the analytical inverse-kinematics method of~\cite{li2021hybrik} using swing-twist decomposition. Similar to~\cite{li2021hybrik}, we use the parametric body model SMPL~\cite{SMPL:2015} to kinematically represent the body motion. SMPL consists of a linear function that takes the pose parameters $\theta \in \mathbb{R}^{24\times3}$ and shape parameters $\beta \in \mathbb{R}^{10}$ as input and produces an articulated triangle mesh $\mathbf{M} \in \mathbb{R}^{6980\times3}$ containing $6890$ vertices. Like in SMPL, we paramaterize joint rotations with the exponential map representation. The method of~\cite{li2021hybrik} uses a trained model to predict the twist component for all body joints. In this work, we initially set the twist to zero, and optimize it as part of the physics optimization as we will explain later in this section. 

Given the 3D body pose in all video frames $t=[0,T]$ represented as local rotations $\theta_t$, we first remove the the high-frequency noise by smoothing the motion using a Butterworth low-pass filter. The exponential map rotation suffers from singularities at $2\pi$ rotations, hence we model global root rotation by separating out global yaw rotation and representing it with per-frame rotation offsets. Specifically\footnote{We apply all mathematical operations on joint rotations including the optimization using quaternion, but leave the conversion for brevity.},
$$\theta^{root}_t = \left(\sum_{\tau=0}^t \Delta \theta^{root,yaw}_\tau\right) * \theta^{root,xy}_t$$
We apply the same smoothing procedure to the global root positions $p^{root}_t$ as well.
Gathering these, we represent the overall motion in generalized coordinates $q_t = \{p^{root}_t, \theta_t\}$.

Although this motion sequence can then be optimized directly, we further model the motion with cubic splines to constrain our motions to be smooth and reduce the dimensionality of our optimization variables. Specifically, we use cubic Hermite splines where the node positions of the spline are initialized to temporally evenly spaced frames covering the whole motion (effectively subsampling it by a factor of 8) and the tangents are initialized according to the rule for Catmull-Rom splines. To compute the full motion sequence, we simply query the spline at the sampling times of the original motion sequence. 
We also optimize the timings $\Delta t_i$ between the spline knots but found its inclusion to have negligible impact on the final results. 

\begin{table}[t]
\centering
\begin{tabular}{p{0.2\linewidth} | p{0.6\linewidth} }
Variable                & Description                                                    \\
\hline
$\beta$                 & Body shape parameter for SMPL model. It is static over time.  \\
$p^{root}_t$            & Root position.                                               \\
$\Delta \theta^{root,yaw}_t$ & Delta axis angle rotation only along the z gravity direction.         \\
$\theta^{joints}_t$ & Axis angle local rotation of joints root for each spline knot.         \\
$\theta^{root,xy}_t$         & XY rotation of root as unnormalized xy quaternion.            \\
$f^c_{t}$           & Scaled contact forces at the $n_c$ contact sites.             \\
$\Delta t_i$            & Delta time to next spline knot. 
\end{tabular}
\vspace{-1mm}
\caption{\small Overview of variables that are directly optimized, their symbol and description. For all of the variables that depend on time, we are actually optimizing the parameters of their respective splines (including tangent values).}
\label{tab:variable}
\vspace{-2mm}
\end{table}

\subsection{Motion Optimization}
In the motion optimization step, we refine the motion by jointly optimizing the body shape $\beta$ and global character poses $\{q_t\}_{t=1:T}$ to match both the pose estimator detections as well as a full-body physics loss term that uses a smooth contact penalty~\cite{mordatch2013animating}. Note that we optimize only one set of shape parameters $\beta$ for the entire sequence, as the identity of the person does not change within a sequence. This stage also optimizes corresponding ground contact forces $f^c_t$ which we parameterize with splines just as we do for the pose.
The total loss function to be optimized in our method combines a physics loss with a pose estimation loss and a smoothness regularization. 
$$L_{total} = L_{pose} + L_{physics} + L_{smooth}$$
We detail each part below.
We evaluate the losses at evenly spaced discrete time points in the motion and average over the entire sequence. 

\subsubsection{Physics Loss}
We now detail the computation of our differentiable physics loss function, given a motion and associated contact forces.

Assume that a temporally evenly spaced sequence of motion frames $\{q_t\}_{1:T}$ and contact forces $\{f^c_t\}_{1:T}$ are given, where $q_t$ and $f^c_t$ represent the generalized coordinates and global contact forces of the body at time $t$. 
The loss function consists of three main parts: 
\begin{equation}
L_{{physics}}(q_t,f^c_t) = L_{{dynamics}} + L_{{contact}} + L_{penetration}
\end{equation}
The dynamics loss penalizes impossible forces.
Rigid body dynamics satisfy the Newton-Euler equations which admits a unique inverse dynamics function mapping motions to the required generalized forces that would give rise to them.
\begin{equation}
f^r_t(q(\cdot)) = M\ddot{q}_t + C\dot{q}_t + g
\end{equation}
The inverse dynamics computation involving mass matrix $M$, centrifugal and coriolis forces $C\dot{q}_t$ and gravity $g$ can be efficiently computed using the Recursive Newton Euler algorithm which exploits the sparsity structure induced by the kinematic tree and we use finite difference approximations for the time derivatives of $q(t)$.
For a thorough tutorial on rigid body dynamics, one can refer to \cite{liu2012quick}. Using $f^r_t$ we can calculate the dynamics loss by comparing it to the actual forces on the character.
\begin{equation}
L_{{dynamics}} = w_{dynamics}||f^r_t - Bf^a_t - J^T f^c_t||^2
\end{equation}
Here $J^T$ maps all the contact forces from the contact points onto the full space and similarly $B$ maps joint actuation $f^a_t$ forces to the full space.
Instead of leaving $f^a_t$ as yet another optimization variable, the optimal value of $B f^a_t$ can be easily chosen by assuming no limits on actuation force.
Practically this means that only residual forces on the root (and other unactuated joints) will be penalized and otherwise it is assumed that any extra acceleration is due to actuation.
Magnitude of joint actuation is implicitly limited by penalizing acceleration of 3d joint positions and rotations, described later.

The humanoid character model is approximated with boxes, cylinders and spheres and differentiably scaled as a function of the skeleton. We detail this in the supplementary. Full-body inertia is accurately accounted for in the inverse dynamics loss and does not make use of centroidal approximations as in prior work~\cite{mordatch2012contact}.
Contact forces are assumed to be exerted only by the feet at 4 different contact points (which we will refer to as end effectors) per foot that lie on the corners of the box approximation to the feet as in~\cite{PhysCapTOG2020} and~\cite{RempeContactDynamics2020}, although more contact locations could be readily added to the current framework.

The contact cost penalizes violation of Signorini's conditions for contact:
\begin{equation}
L_{contact} = \sum_i^{n_c} c_{t,i}\left({w_e||e_{t,i}||}^2+{w_{\dot{e}}||\dot{e}_{t,i}||}^2\right)
\end{equation}
Here, $e_{t,i} \in \mathbb{R}^3$ is the minimum displacement between the i\textsuperscript{th} end effector position and the contact surface and its time derivative $\dot{e}_{t,i}$ is also included to prevent slip.
They are penalized in proportion to the contact variable $c_{t,i}$ related to the contact force.
The contact variable represents the degree to which a contact is present at that time step.
It ranges from 0 to 1 and is obtained through a soft step function of the contact force magnitude as:
\begin{equation}
c_{t,i} = \frac{1}{2}(\tanh(k_1 ||f^c_{t,i}|| - k_2)+1)
\end{equation}
The contact variable is a monotonically increasing function of the contact force.
Furthermore, it saturates for large values of $f^c$. 
This can be seen as a soft-relaxation of the hard step function in the complementarity condition.
Intuitively, optimality is reached by bringing the contact force to zero and/or the contact distance to zero.
Slipping and rolling contacts are not considered in this work.

Without further restrictions the contact objective only penalizes violation of the Signorini conditions when it specifically chooses to apply contact force.
As such the method can generate motions with penetrating objects without contact force.
To avoid this, a separate term is used to explicitly penalize interpenetration:
\begin{equation}
L_{penetration} = w_{pen}\sum_i^{n_c} \max(\{d_{t,i}+k_{margin},0\})^2
\end{equation}
Here, $d_{t,i}$ is the signed distance of the contact surface at the i\textsuperscript{th} end effector which is negative if it is penetrating.

\subsubsection{Pose Estimation Loss}

The pose fitting loss $L_{pose}$ we use is common in human shape estimation~\cite{Bogo:ECCV:2016}.
$L_{pose}$ is evaluated per frame and summed. It measures the motion error in terms of local 3d keypoints deviation, global camera projected 2d keypoint deviations, log probability of the motion under a pose prior and deviation of the SMPL body shape from the mean body shape.

We also use a kinematic acceleration penalty to ensure our motions are smooth.
\begin{equation}
L_{smooth} = \frac{1}{n_{joints}}(w_{\ddot{\theta}}||\ddot{\theta}_t||^2 + w_{\ddot{p}}||\ddot{p}_t||^2)
\end{equation}
Here $\ddot{p}_t$ is the global linear acceleration of the joints. 

All of our loss terms have tuned weights which are detailed in the supplementary along with additional loss details. Although our method is not overly sensitive to this tuning, it is important to have a good balance between weighing $L_{dynamics}$ and $L_{contact}$.
Outside of that, the balance between $L_{physics}$ and $L_{pose}$ was loosely tuned such that $L_{pose}$ does not deviate much from a purely kinematic optimization.
\subsubsection{Implementation Details}

We implement our full pipeline in PyTorch and use an off-the-shelf implementation of the LBFGS optimizer~\cite{NoceWrig06} with a history size of 100, base step size of 1.0 and Armijo-Wolfe line search.
The optimization is run in 2 stages totalling 750 iterations. 
First 250 iterations of kinematic optimization is performed where the only difference is that $L_{physics}$ loss is disabled, then 500 iterations of physics optimization is performed with $L_{physics}$ enabled. 
The LBFGS memory is cleared between the 2 stages.

\subsection{Generative Model}
Once our motion has been optimized we can use it like a standard motion capture dataset.

In particular, we demonstrate that it can be used to train motion synthesis model that are typically only trained on mocap datasets.
We follow prior work in generative human motion synthesis and adopt the state of the art Diversifying Latent Flows (DLow) method~\cite{yuan2020dlow}. DLow uses a standard recurrent conditional VAE (CVAE) with a GRU encoder, auto-regressive decoder architecture to predict future motion given a short clip of past motion as context. Additionally it uses a learned post-hoc sampling strategy that optimizes directly for both best-of-1 accuracy and diversity of a finite number set of future motion predictions.

DLow takes as input and produces as output a sequence of root relative 3d keypoint positions and root velocities.

\section{Experimental Results}
In this section we evaluate our method and compare to previous work. We split our evaluations into the two stages of our pipeline.
We first provide experimental details of our evaluation setting~\ref{sec:experimental_setting}. Next, we evaluate our physics refinement step for pose estimation and compare against state of the art physics-based approach PhysCap~\cite{PhysCapTOG2020} and pose estimators HMR~\cite{hmrKanazawa18}, HMMR~\cite{humanMotionKanazawa19}, and VIBE~\cite{kocabas2019vibe}. Finally, we demonstrate the benefits of using our physics optimization correction in terms of downstream performance on motion synthesis.

\begin{table*}[t!]
\centering
\begin{tabular}{l|lllll|l}
                     & HMR~\cite{hmrKanazawa18}   & HMMR~\cite{humanMotionKanazawa19}  & PhysCap~\cite{PhysCapTOG2020} & Ours (kin) & Ours (dyn)    & VIBE*                         \\ \hline
no Procrustes MPJPE ($\downarrow$)  & 78.9  & 79.4  & 97.4    & 73.6       & \textbf{68.1} & 65.6                         \\
global root position ($\downarrow$) & 204.2 & 231.1 & 182.6   & 148.2      & \textbf{85.1} & -                            \\
$e_{smooth}$  ($\downarrow$)       & 11.2  & 6.8   & 7.2     & 5.42       & \textbf{4.0}  & -                            \\
$\sigma_{smooth}$ ($\downarrow$)   & 12.7  & 5.9   & 6.9     & \textbf{1.06}       & 1.3 & -  \\
\hline
\end{tabular}
\vspace{-1mm}
\caption{Comparison of pose estimation accuracy and quality metrics for our method with physics (dyn) and without physics (kin) along with competitive pose estimator baselines. 
All errors are measured in millimeters.
VIBE~\cite{kocabas2019vibe} is a strong oracle method that uses the large-scale AMASS~\cite{AMASS:ICCV:2019} motion capture dataset for training. Note that as PhysCap~\cite{PhysCapTOG2020} and the other baselines operate at 25fps, we downsample our 50fps motion for making a direct comparison. }
\label{tab:pose_est_acc}
\end{table*}

\subsection{Dataset and Experimental Setting}
\label{sec:experimental_setting}
We use the large scale Human3.6M dataset for our evaluations (and additional comparison to~\cite{RempeContactDynamics2020} on HumanEva~\cite{Sigal_2010} is provided in the supplementary). Motions were recorded from 4 cameras and a motion capture system was used to produce accurate annotations for the character.
We use subjects 9 and 11 which form the standard validation set and use the same motions as PhysCap~\cite{PhysCapTOG2020}. Specifically, these motions do not include interactions with the chair object or lying/sitting motions. They are: \textit{directions, discussions, greeting, posing, purchases, taking photos, waiting, walking, walking dog and walking together}. 

\subsection{Physics-corrected Pose Estimation}
Through our evaluation we want to answer the following questions: {\bf 1)} Does our proposed physics loss improve the accuracy of pose estimation?, {\bf 2)} Does it improve the physical plausibility of pose estimation?, and {\bf 3)} How does our method compare against other physics/temporal pose estimation methods?

As we do not have access to the DeepCap dataset~\cite{habermann2020deepcap}, we evaluate our method on the large scale Human3.6m dataset. We split motions into even chunks such that they are below 2000 frames (40 seconds). Most motions can be processed in one or two chunks, but a few motions require three chunks. Optimization completes in ~3-4 minutes for a chunk of length 40 seconds.

\vspace{-2mm}
\paragraph{Baselines.} To address the first 2 points, we introduce a kinematic optimization baseline, which is equivalent to our method in all aspects, except that $L_{physics}$ is not included in the total loss for optimization (and consequently the end effector forces are also not included in the optimization variables). We also compare against HMMR~\cite{humanMotionKanazawa19}, which is a kinematic 3D mesh and pose prediction model from videos of human motions in the wild. We further compare to its predecessor, HMR~\cite{hmrKanazawa18}, which performs a similar function given a single RGB image, as opposed to a video. Our third baseline is PhysCap~\cite{PhysCapTOG2020}, a physics-based 3D pose prediction model from monocular video that includes contact modeling and  minimizes  foot-to-floor  penetration, unlike other similar methods. We also compare against VIBE~\cite{kocabas2019vibe}, a strong oracle that predicts both pose and body shape, but which has been trained on the large-scale AMASS~\cite{AMASS:ICCV:2019} motion capture dataset. Similarly to HMR and HMMR, VIBE relies on an adversarial objectives that discriminates between mocap motion and predicted motion.  
 
\vspace{-2mm}
\paragraph{Evaluation metrics.} We adopt evaluation metrics outlined in PhysCap~\cite{PhysCapTOG2020}.
Following standard practice, we measure mean per joint position error (MPJPE) on the 15 joint reduced skeleton and the mean global root position error.
The $e_{smooth}$ loss is also introduced in PhysCap and we report it as well. 
It measures the difference in 3d keypoint velocity magnitude between the ground truth motion and the predicted motion which illustrates the amount of jittering present in the motion and is computed as follows:
\begin{align}
    \hat{Jit} &= ||\hat{p}_{t}-\hat{p}_{t-1}|| \\
    Jit^{GT} &= ||p^{GT}_{t}-p^{GT}_{t-1}|| \\
    e_{smooth} &= \sum_{t}^T \sum_{joints}||p^{GT}_{t}-p^{GT}_{t-1}||
\end{align}

Pose estimators that do not make use of physics losses often violate the conditions of static contact.
We create metrics based on the foot joint that directly aim to measure this. 
Contact condition violation occurs in two ways which we design metrics for to test.
To evaluate foot floating artefacts, we compare foot global $z$ position error ($e_{foot,z}$) on ground truth: 
\begin{equation}
e_{foot,z} = mean(|\hat{p}_{foot,z} - p^{GT}_{foot,z}|)
\end{equation}
To evaluate foot sliding artefacts, we compare foot global $xy$ velocity error ($e_{foot,vxy}$) with respect to ground truth.
\begin{equation}
e_{foot,vxy} = mean(||\Delta_t\hat{p}_{foot,xy} - \Delta_t p^{GT}_{foot,xy}||)
\end{equation}

\paragraph{Results.} We detail our pose estimation accuracy results in Table~\ref{tab:pose_est_acc}. Our method greatly outperforms PhysCap~\cite{PhysCapTOG2020} on root-aligned mean joint position error without procrustes alignment. In fact, our method approaches learning-based video pose estimation methods that leverage the large scale AMASS motion capture datasets~\cite{AMASS:ICCV:2019} to form a motion prior. Our kinematic motion baseline is competitive with HMR~\cite{hmrKanazawa18} and HMMR~\cite{humanMotionKanazawa19} on its own, demonstrating the power of optimization-based pose estimation. 

\begin{figure}
    \centering
    \includegraphics[width=0.32\textwidth]{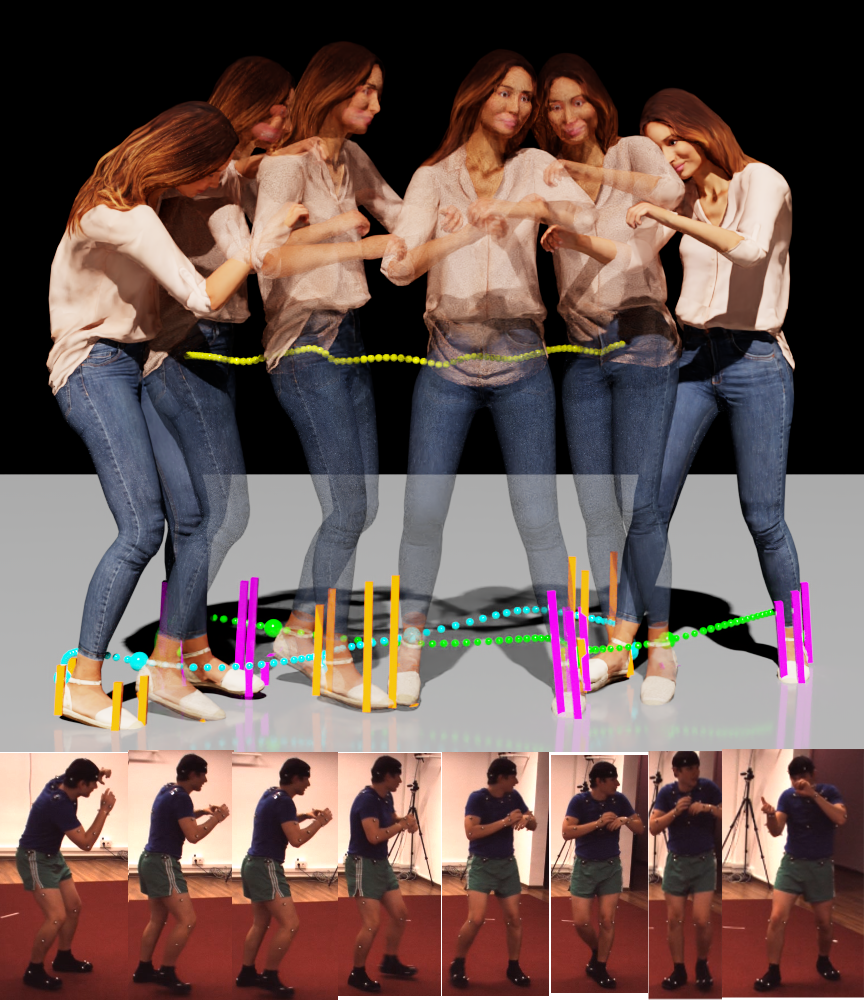}
    \vspace{-2mm}
    \caption{\small{\textbf{Optimization result on video.} Here we show a photo snapping motion produced by our framework, video frames from the input motion are included below.
    }}
    \label{fig:collage}
     \vspace{-2mm}
\end{figure}

Furthermore, we greatly improve in terms of global root position estimation as well.
We attribute this to the fact that we optimize motion and body shape jointly along with our contact-aware physics loss.
Thereby the movement of 2d joint detections over time can help estimation of the bone lengths as can be seen Fig~\ref{fig:collage}, instead of taking the initial average bone lengths without further refinement as done in PhysCap~\cite{PhysCapTOG2020}. This shows that relying on temporally learned pose estimators alone to recover global scale and bone lengths is suboptimal. PhysCap does not have a direct mechanism to allow these bone lengths to be optimized with respect to contact aware losses as they optimize in separate iterative stages. However, we have a single differentiable objective that is jointly optimized with all variables including shape parameters. 

The addition of the physics loss makes a noticable improvement in terms of MPJPE and a very large improvement on global root position and $e_{smooth}$. The large improvement in the joint speed error $e_{smooth}$ is immediately visible in videos of the two approaches which are included in supplementary.
For our kinematic baseline, without guidance on when contacts are formed and broken, feet joints of the character can often slide side to side during contact and without enforcement of the Newton Euler equations, the root of the character can freely move without limitations and often slide side to side during fast walking phases. 

The major contributor to the difference in global root position error is due to depth ambiguity.
Whereas, the kinematic baseline can only form a rough approximation to depth using body priors and motion cues, our physics loss enforces contact with the ground plane directly, greatly improving depth estimation. We further find the benefits of including the physics loss on the physical plausibility of contacts through our custom metrics outlined in Table~\ref{tab:contact_metrics}. Specifically, the physics loss reduces foot tangent velocity error by more than $40\%$ and height error by $80\%$.

\begin{table}[]
\centering
\begin{tabular}{l|ll}
          & $e_{foot,vxy}$ ($\downarrow$) & $e_{foot,z}$ ($\downarrow$)\\ \hline
Ours (kin) &  4.65          & 95.7       \\
\textbf{Ours (dyn)} & \textbf{2.71}           & \textbf{18.9}         \\ \hline
\end{tabular}
\vspace{-2mm}
\caption{\small Ablation comparison of contact-sensitive metrics, foot tangential velocity error ($e_{foot,vxy}$) and foot global height error ($e_{foot,z}$) with and without physics loss.}
\label{tab:contact_metrics}
\vspace{-2mm}
\end{table}

\begin{figure}
    \centering
    \includegraphics[height=3.5cm]{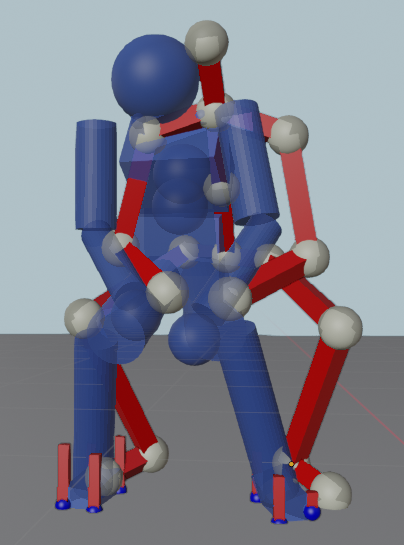}
    \includegraphics[height=3.5cm]{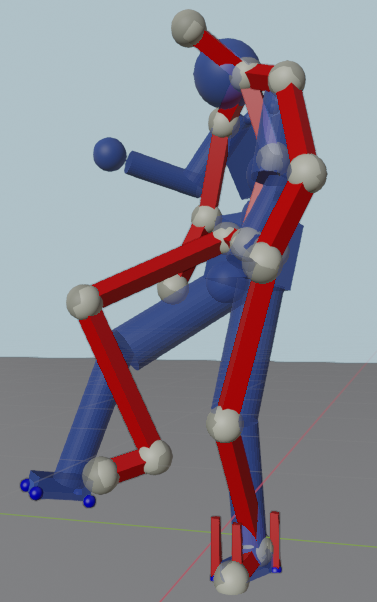}
    \caption{\small{\textbf{Failure cases.} Even when the mocap reconstruction error is quite high, the motion output of our method still evaluates consistently low under our physics loss and visually looks physically plausible. These failure cases 
    are selected from the worst performing frames in terms of mocap reconstruction.}}
    \label{fig:fail}
\end{figure}

\vspace{-2mm}
\paragraph{Qualitative results.}
Measuring the quality of motion capture is difficult and quantitative metrics do not always paint the full picture.
We include qualitative examples of our output as composited renders. We also show representative failure cases from the most inaccurate frames of our predictions in Fig.~\ref{fig:fail}.

\begin{figure}
    \centering
    \includegraphics[height=3.5cm]{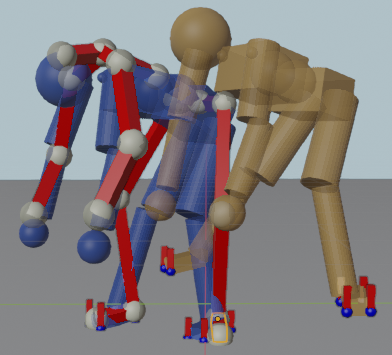}
    \includegraphics[height=3.5cm]{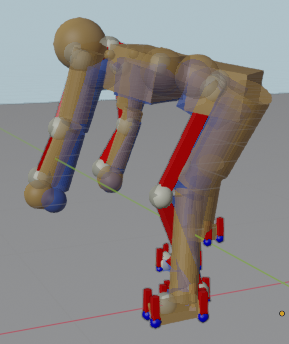}
    \vspace{-2mm}
    \caption{\small{\textbf{Pose estimation result.} In light orange is the motion initialization for our optimization, in blue is the final output of our method overlayed on the red skeleton which is ground truth joints. In the camera view on the right, the initial pose looks plausible, but is refined drastically as the body shape is optimized by our method as seen on the side view shown on the left. 
    }}
    \label{fig:singletasks}
     \vspace{-2mm}
\end{figure}

 Many of the largest error cases occur near crouching motions. Here we are mainly limited by our geometric character approximation. The box geometry of our character does not capture the true underlying foot geometry. Our model is unable to represent significant foot flexion. However, we note that the character pose is still stable and the contact of our end effectors still engaged with the ground plane. 

We also note that estimated contacts are also realistic. In fact, even though we do not use a contact detection network, we are still able to estimate realistic forces from only video input. Figure~\ref{fig:collage} qualitatively demonstrates estimated ground contact forces during a typical walking gait.

\subsection{Motion Synthesis from Video}

Here we show results for our combined framework that trains a motion generative model from video.
Through our evaluation we want to answer the following questions: {\bf 1)} When trained with only our pose estimation-generated data, can we learn high quality motion synthesis models?, {\bf 2)} How much does our physics loss in the pose estimation step improve the performance of the down-stream motion synthesis model?

To address these questions, we train the same DLow~\cite{yuan2020dlow} model with 3 different training datasets.
DLow(GT) is the oracle model that trains with actual mocap data.
DLow(PE-dyn) is our proposed method that uses the physics corrected pose estimation results from the previous stage. 
DLow(PE-kin) is a baseline that uses the kinematically-optimized pose estimation results from the previous stage and is used to ablate the benefit of using physics loss. We also include the results of the standard VAEs trained in the different fashions without DLow sampling.

We keep experimental settings identical between the 3 and the only varying factor is the input training data.
Ultimately we evaluate the trained motion synthesis models on the ground truth validation set. As the Human3.6M~\cite{h36m_pami} dataset contains multi-view cameras for each motion, we only make use of videos from the first camera to generate our datasets which simulates a monocular RGB video setting.
We follow a similar evaluation protocol to DLow~\cite{yuan2020dlow} and compare against it.

Additionally, as we limit our pose estimator to the validation set of Human3.6m, we only train the motion synthesis model with motions from the two characters, (S9 and S11).
Therefore we split the motions from S9 and S11 evenly into a training and evaluation set for the motion synthesis model. Specifically, every motion named \textit{'[Action] 1'} is used for validation leaving, one other motion of the same action type in the training set.

Apart from this, we use the exact same experimental settings as in DLow. Given a context of 0.5 seconds, DLow predicts the future 2 seconds of the motion. All motions are sampled at the original 50FPS of the mocap.
We use DLow with 10 output motion modes.

\vspace{-3mm}
\paragraph{Evaluation metrics.} We report standard  metrics used in motion synthesis.
The two distinct objectives of motion synthesis is to generate diverse, yet accurate motions.
Accuracy is measured on the 15 joint skeleton model, average distance error (ADE) measures average root-aligned joint position error averaged over the predicted future motion sequence and final distance error (FDE) is the same, but measured only at the final frame of the predicted motion, which emphasizes longer term accuracy. Both metrics are in meters. Diversity is measured by average pairwise distance (APD). Given the set of samples produced by the motion synthesizer this gives the average L2 distance between all pairs of motion samples.

\vspace{-3mm}
\paragraph{Results.} We tabulate the evaluation of our models in Table~\ref{tab:mgen_acc}. The cVAE is the VAE that forms the backbone for the DLow method. As expected we do not match the quality of the oracle model which uses ground truth motion capture data.
However, we are very competitive with this oracle.
DLow trained using our physics corrected input (PE-dyn) is only worse in average joint distance by 16.9\%, in final distance error by 11.0\% and in average motion diversity by 10.3\%.
For both the DLow model and the cVAE model, adding physics loss to the correction step for generating training data consistently improves all evaluated metrics. 

\vspace{-3mm}
\paragraph{Qualitative results.} Please see the supplementary for videos and visualizations of the motions produced from our trained motion model.

\begin{table}[t!]
\centering
\begin{tabular}{lllll}
\hline
             & Diversity ($\uparrow$)     & ADE  ($\downarrow$)           & FDE ($\downarrow$)            \\ \hline
DLow(PE-kin) &  10.53    &  0.590  &  0.698         \\
DLow(PE-dyn) & \textbf{10.96}         & \textbf{0.573}           & \textbf{0.685} \\ \hline
DLow(GT)*    & 12.22 & 0.490  & 0.617  \\ \hline
cVAE(PE-kin) & 7.419          & 0.639           & 0.756           \\
cVAE(PE-dyn) & \textbf{7.413}          & \textbf{0.612}           & \textbf{0.738}           \\ \hline
cVAE(GT)*    & 6.801 & 0.5617 & 0.706 \\
ERD(GT)*     & 0              & 0.722           & 0.969 \\
\hline
 \end{tabular}
\vspace{-2mm}
\caption{\small Comparison of motion synthesis diversity and accuracy between motion synthesis models with different training data. Note that the errors are measured in meters as we stick to the convention in motion synthesis works. 
The (GT)* denotes that the method was trained with ground truth mocap data, not estimated from video and should be understood to be an oracle baseline.
PE-dyn is using our physics corrected pose estimation dataset and PE-kin is ablating away the physics loss in the physics correction.
}
\label{tab:mgen_acc}
\end{table}

\vspace{-2mm}

\section{Conclusion}
In this paper, we introduced a new framework for training motion synthesis models from raw video pose estimations without making use of motion capture data. Our framework refines noisy pose estimates by enforcing physics constraints through contact invariant optimization, including computation of contact forces.
We then train a time-series generative model on the refined poses, synthesizing both future motion and contact forces. Our results demonstrated significant performance boosts in both, pose-estimation via our physics-based refinement, and motion synthesis results from video. We hope that our work will lead to more scaleable human motion synthesis by leveraging large online video resources.

{\small
\bibliographystyle{ieee_fullname}
\bibliography{egbib}
}

\newpage 
\begin{center}
\textbf{{\Large Supplementary Material}} 
\end{center}
\appendix 

\section{Detailed Loss Formulation}
Here we detail the losses used in our optimization method.
To obtain the physics losses we first must obtain velocity and acceleration of our character. We use a finite difference scheme that corresponds to implicit integration.
$$\dot{q}_t \approx (q_{t+1}-q_{t})/\Delta t$$
$$\ddot{q}_t \approx (\dot{q}_{t+1}-\dot{q}_{t})/\Delta t$$

For the contact loss, we compute the contact variables $c_{t,i}$ using $k_1$ and $k_2$ parameters. Here $k_1$ controls the stiffness of the contact. The higher it is, the closer the soft contact loss approaches a hard step function corresponding to contact complementarity constraint. Then $k_2$ is simply an offset that ensures that $c_{t,i}=0$ when $f_{t,i}=0$.
We employ 2 additional penalties that keep the contact forces physical.
First we penalize contact forces from violating the friction cone constraint.
We set the friction constant $\mu=1.0$ (which is a generous overestimate representing rubber on rubber contact) and calculate the deviation from this.
$$L_{friction}(t) = w_{\mu}\sum_i^{n_c} \max\left(\frac{||{f^c_i(t)}_\parallel||^2}{||{f^c_i(t)}_\perp||^2} -\mu,0\right)$$
Here $\parallel$ indicates the component of force tangential to the contact surface and $\perp$ the normal force.
The second one is to prevent overly excessive contact forces that are unreasonable for natural human motion.
We set this to $8$ times the force required for an evenly balanced standing motion.
Since there are $8$ foot contact points, each contact point would hence be restricted to exert no more than the whole body weight on its own.
This means each foot can generate total contact force of $4$ times body weight which is similar to highly dynamic dance motions.

The pose loss is composed of thee parts.
$$L_{pose} = L_{prior} + L_{pose2d} + L_{pose3d}$$
$L_{prior}$ is the per-frame SMPL prior  and is the log prob. of a Gaussian Mixture Model over pose and L2 regularization of the body shape parameter $\beta$.
$$ L_{prior} =  w_\beta ||\beta||_2^2 + w_{GMM} \sum_t \log p_{GMM}(\bm{\theta^{joints}}_t)$$
$L_{pose2d}$ is error in pixel space, re-projecting our motion using true camera projection matrix $P$ and uses robust loss $\rho$.
$$ L_{pose2d} = w_{2d} \rho(Pp - x^{pe,2D}) $$
$L_{pose3d}$ measures local keypoint 3d error where global root position is subtracted out to obtain relative keypoint positions $p_{rel}$. Here $R$ is the camera extrinsic rotation.
$$ L_{pose3d} = w_{3d} ||Rp_{rel}-sp^{pe}_{rel}||^2 + w_{scale} (s-1)^2$$
Since scale of the original 3d pose estimation is inherently ambiguous, the scale parameter $s$ is jointly optimized with the motion which accounts for this ambiguity. The actual scale of the character in our optimization will be adjusted through the $\beta$ shape parameters and informed through contact geometry and motion (scale typically does not diverge too much from 1).
In our case the pose estimator we use also emits a score representing the confidence in its estimation (ranging from 0 to 1). In this case, we also weigh the pose estimation losses per joint by this confidence.

\begin{table}[]
\centering
\begin{tabular}{|l|l|}
\hline
Name                & Value    \\ \hline
$w_{dynamics}$      & 50       \\ \hline
$w_e$               & 200      \\ \hline
$w_{\dot{e}}$       & 50       \\ \hline
$k_1$               & 10       \\ \hline
$w_{\mu}$            & 1        \\ \hline
$w_{pen}$           & 100      \\ \hline
$w_{2d}$            & 1e-3   \\ \hline
$w_{3d}$            & 0.5      \\ \hline
$w_{scale}$         & 1e-3   \\ \hline
$w_{\beta}$         & 5e-3   \\ \hline
$w_{GMM}$           & 2.5e-3 \\ \hline
$w_{\ddot{p}}$      & 0.15     \\ \hline
$w_{\ddot{\theta}}$ & 1e-4   \\ \hline
\end{tabular}

\caption{Table of constants used and their values.}
\label{tab:constants}
\end{table}

We give weights for each of the losses in Table~\ref{tab:constants}
\section{Rigid Body Human Body Model}
\begin{figure}
    \centering
    \includegraphics[height=5cm]{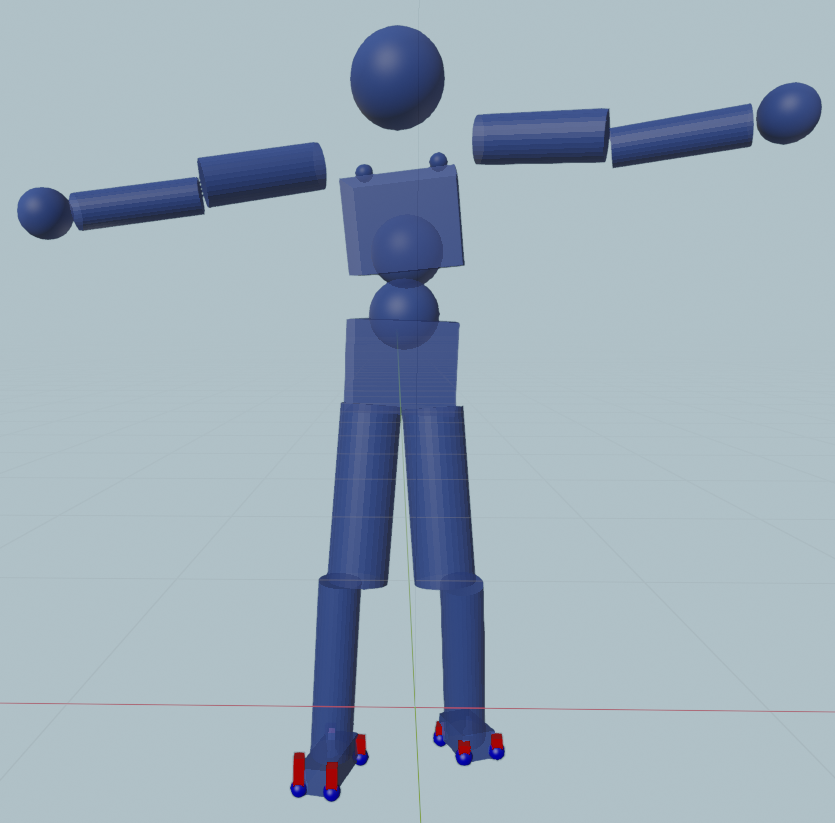}
    \caption{\small{\textbf{Body Model.} Example geometry of our human body model.}}
    \label{fig:body_def}
\end{figure}

We construct the body model out of geometric primitives as shown in Figure~\ref{fig:body_def}. 
Mass and inertial properties are calculated assuming constant density of $1000kg/m^3$.

Sizes of primitives are heuristically set and are (differentiably) scaled in proportion to the lengths of the corresponding bones of the skeleton resulting from the SMPL body shape params $\beta$.
Specifically, we scale box and sphere primitives corresponding to foot, torso and head uniformly in all 3d dimensions in proportion to the distance to the next child joint, whereas we scale the cylinders representing limbs only in the length-wise direction and maintain a constant thickness.
Admittedly, this does not fully capture variation in human body shapes, as it does not distinguish between characters with similar skeletons that differ in body mass. 

\begin{table}[]
\centering
\begin{tabular}{lccc}
\hline
\textbf{Method}         & \textbf{Feet} & \textbf{Body}  & \textbf{Body-Align 1} \\ \hline
~\cite{RempeContactDynamics2020} Physics (MTC)    & 508.7         & 499.8          & 421.9    \\
~\cite{RempeContactDynamics2020} Physics (Our PE) & 345.2         & 382.0          & 310.8   \\ \hline
Ours (Kinematic)        & 251.0         & 190.1          & \textbf{114.6}   \\
Ours (Physics)          & \textbf{82.4} & \textbf{101.1} & 156.0 \\ \hline
\end{tabular}
\caption{Comparison with~\cite{RempeContactDynamics2020} on HumanEva dataset. Errors are mean over time and measured in millimeters.}
\label{tab:humaneva}
\end{table}

\section{Additional Comparison}

We adopt the same experimental setting presented in~\cite{RempeContactDynamics2020}. Specifically we evaluate on the same 15 short (2 second) sequences extracted from the walking clips in the HumanEva~\cite{Sigal_2010} dataset.
Using the camera extrinsics given in the dataset, we found that the ground had a slight z offset in the clips. We estimated a conservative 6cm offset for all clips and applied our method with this elevated ground plane. 
In~\cite{RempeContactDynamics2020}, MTC is used as the pose estimator for the initial motion.
For fairer comparison, we also adapt a version that uses the outputs of our pose estimator (Our PE) instead.
Note that unlike our method,~\cite{RempeContactDynamics2020} does not optimize the shape of the body during optimization which can lead to large errors especially in the depth of the root.

We present our results using the metrics reported in~\cite{RempeContactDynamics2020} in Table~\ref{tab:humaneva}.
Here ``Feet" measures the global position error of the 2 feet joints. This especially highlights foot floating and sliding artifacts.
The ``Body" metric measures whole body global position error.

The metric ``Body-Align 1" measures the average error between the poses after aligning the 1st frame root position.
This metric is not very robust as it is quite sensitive to the root position on the first frame.
For example, if one motion started with an error in the root position on the first frame but later in the motion recovers from this error, it will be penalized for the rest of the motion as the first frame is erroneously compensated for.

Our method greatly outperforms~\cite{RempeContactDynamics2020} in every pose accuracy metric.

\end{document}